\documentclass{esannV2}

\usepackage[dvips]{graphicx}
\usepackage[latin1]{inputenc}
\usepackage{amssymb,amsmath}
\usepackage{microtype}

\title{Mental Model Management: An Operator-Based Framework for LLM Memory}

\author{
Oliver Kramer\\
Computational Intelligence Lab\\
Department of Computer Science\\
University of Oldenburg, Germany\\
oliver.kramer@uni-oldenburg.de
}

\begin{document}

\maketitle

\begin{abstract}
Large language models process large amounts of information but usually
lack an explicit mechanism for maintaining compact and evolving
conceptual representations. We introduce Mental Model Management (3M),
a framework in which knowledge is represented as mental models consisting
of compact chunks. Rather than accumulating text passages, 3M continuously
integrates new information into an existing conceptual representation.
A set of operators extracts knowledge, retrieves relevant models, adds
and updates chunks, reorganizes representations, detects inconsistencies,
and derives new knowledge. We describe the main 3M operators and
illustrate each operation using Evolution Strategies as a running example.
\end{abstract}
\section{Introduction}

Large language models (LLMs) can extract, combine, and reason about
information in their context, but their knowledge processing remains
strongly coupled to model parameters and the current context.
Retrieval-augmented generation (RAG) extends this context with external
information \cite{lewis2020rag,karpukhin2020dpr}. Long-term memory has
similarly become an important component of LLM-based agents. Systems such
as MemGPT \cite{packer2023memgpt}, MemoryBank
\cite{zhong2023memorybank}, Generative Agents
\cite{park2023generative}, and A-MEM \cite{xu2025amem} provide mechanisms
for persistent storage, retrieval, reflection, updating, and increasingly
dynamic memory organization.

These approaches substantially extend LLM capabilities. However, they do
not by themselves provide an explicit vocabulary for transforming an
evolving conceptual representation. Information may be redundant,
outdated, disconnected, or contradictory, while general principles may
remain implicit across multiple observations.

Mental Model Management (3M) addresses this problem by representing
persistent knowledge as an evolving collection of conceptual models and
defining explicit operators for their transformation. The term
\emph{mental model} has a long history in cognitive science as a
representation supporting understanding and inference
\cite{johnsonlaird1983mental}. In 3M, a mental model has a more specific
computational meaning: it represents a concept through a small collection
of compact and revisable knowledge chunks.

Incoming information is not simply appended to memory. Instead, relevant
models can be extended, updated, merged, split, connected, compressed, or
corrected. Cognitive operations can additionally derive, generalize, or
abstract knowledge. This perspective is related to belief revision
\cite{alchourron1985logic}, but extends the operator view toward memory
organization and cognitive transformations for LLM-based systems.

Many individual mechanisms underlying 3M have precedents in retrieval,
agent memory, knowledge representation, and cognitive architectures. The
contribution is therefore not that these operations are individually new.
Rather, 3M provides a unified framework in which concept-centered mental
models are transformed through an explicit vocabulary of memory and
cognitive operations. Importantly, derived knowledge, abstractions,
repaired conflicts, and identified gaps can themselves become persistent
parts of the evolving memory.

The following sections introduce the 3M operators. Evolution Strategies
(ES) serve as a running example because they contain closely interacting
concepts such as mutation, selection, step-size adaptation, and covariance
adaptation \cite{hansen2016cma}.

\section{Mental Models}

\subsection{Definition}

The basic unit of 3M is a mental model $M$. A model is associated with a
concept and contains a set of compact chunks,
\begin{equation}
M = \{c_1,c_2,\ldots,c_k\},
\end{equation}
where each $c_i$ represents one relevant unit of knowledge. A chunk may
encode a fact, relation, rule, observation, procedure, hypothesis, or
derived principle. Chunks should be sufficiently short to be individually
retrieved and modified, but sufficiently expressive to represent useful
knowledge.

For example, a mental model for mutation strength may contain
\begin{verbatim}
## Mutation Strength
Description: Controls the scale of mutations in an ES.

- Mutation strength sigma controls mutation scale.
- Small sigma favors local search.
- Large sigma produces larger exploratory steps.
\end{verbatim}
The model is deliberately different from a textual summary. A summary is
normally a compressed representation of one particular document. A mental
model instead integrates information about a concept across potentially
many documents and interactions. It therefore persists while its
individual chunks can change.

A 3M memory can be written as a collection
\begin{equation}
{\cal M} = \{M_1,M_2,\ldots,M_n\}.
\end{equation}
The purpose of the operators introduced below is to transform
${\cal M}$ as new information becomes available.

\subsection{Relation to Existing LLM Memory}

Retrieval-augmented generation typically stores documents or document
chunks and retrieves passages relevant to a query
\cite{lewis2020rag,karpukhin2020dpr}. Graph-based extensions can introduce
additional structure by extracting entities and relations and organizing
information into graph representations \cite{edge2024graphrag}.
Such approaches provide powerful mechanisms for locating external
information, but their primary objective remains supplying relevant
context to the language model.

Persistent agent-memory systems address a related problem. MemGPT manages
information across memory tiers \cite{packer2023memgpt}; MemoryBank
supports long-term storage, retrieval, updating, and forgetting
\cite{zhong2023memorybank}; and Generative Agents combine retrieval with
reflection over accumulated experiences \cite{park2023generative}.
A-MEM goes further toward dynamically evolving memory organization by
creating interconnected memory networks and updating existing
representations when new memories are introduced \cite{xu2025amem}.
Recent surveys consequently identify memory representation, retrieval,
updating, reflection, and consolidation as central problems for
LLM-based agents \cite{zhang2024memorysurvey}.

Mental Model Management focuses specifically on the transformation of the
persistent conceptual representation. Its basic object is a concept rather
than a source passage. Multiple passages can contribute to the same mental
model, while redundant information can disappear, conflicting information
can trigger repair, and derived principles can become explicit parts of
memory.

For example, ten papers may contain different descriptions of mutation
strength. A passage-oriented memory may retain all ten descriptions. A
mental-model memory may instead maintain a single evolving representation:
\begin{verbatim}
## Mutation Strength
- sigma controls mutation scale.
- Useful sigma depends on the search state.
- sigma can be adapted from search feedback.
- sigma can also be encoded and evolved as a
  strategy parameter.
\end{verbatim}
The original documents can still be retained as evidence and provenance,
but they are separated from the compact conceptual representation used
for reasoning. Retrieval and 3M are therefore complementary: retrieval
determines which external information becomes available, while 3M
determines how that information changes persistent conceptual memory.

\subsection{Operator Organization}

The operators have different roles in the management of mental models.
Some primarily control the flow of information between external text and
memory, while others maintain the internal representation or perform
cognitive transformations.

A useful classification is
\begin{center}
\begin{tabular}{ll}
\hline
\textbf{Function} & \textbf{Operators} \\
\hline
Knowledge acquisition &
Extract, Retrieve \\
Memory management &
Add, Update, Merge, Split, Connect, Compress, Prune \\
Cognitive processing &
Conflict Detection, Conflict Repair, Generalize, \\
& Specialize, Abstract, Infer, Analogy, Find Gap, Verify \\
\hline
\end{tabular}
\end{center}
The distinction is not strict. For example, compression is both a memory
operation and a cognitive operation because constructing a compact chunk
may require understanding the common meaning of several existing chunks.
Similarly, verification can result in an update or deletion of stored
knowledge.
The operators should therefore be understood as a vocabulary of possible
transformations rather than as a rigid processing pipeline.

\section{Knowledge Acquisition}

\subsection{Extract}

Extraction is the interface between unstructured information and mental
models. Given a text $T$, the \emph{extract} operator first identifies
concepts that are important enough to be represented and subsequently
extracts compact chunks associated with these concepts. It therefore
performs both conceptual decomposition and knowledge extraction.

Extraction should be selective. Not every sentence needs to become a
chunk, and not every noun phrase needs to become a mental model. The
objective is to identify information that contributes to understanding
the domain. Details that merely repeat existing information or have little
conceptual relevance can be ignored.

Consider the input:
\begin{quote}
Evolution Strategies are stochastic optimization methods. Candidate
solutions are modified by mutation and selected according to fitness.
The mutation strength $\sigma$ determines the size of mutations.
\end{quote}
The manager may identify \emph{Evolution Strategy}, \emph{Selection},
and \emph{Mutation Strength} as important concepts and produce
\begin{verbatim}
## Evolution Strategy
- ES is a stochastic optimization method.
- Candidate solutions are generated by mutation.
- Selection favors candidates with better fitness.

## Mutation Strength
- Mutation strength sigma controls mutation scale.
\end{verbatim}
Extraction is therefore not intended to reproduce the input. Its output
is a candidate conceptual representation that can subsequently be
integrated into existing memory.

\subsection{Retrieve}

Before modifying memory, the manager should determine whether relevant
knowledge already exists. The \emph{retrieve} operator searches the
mental-model collection for models related to the current text, question,
or reasoning problem. Retrieval can be based on model names, textual
descriptions, embeddings, relations, or combinations of these signals.

This operation is important because new information should normally be
interpreted relative to existing knowledge. Without retrieval, an LLM
could repeatedly create new models for concepts that are already present.
Retrieval therefore provides the context required by later operations
such as update, merge, conflict detection, and inference.

For example, given the query
\begin{quote}
How does step-size adaptation work in Evolution Strategies?
\end{quote}
the system may retrieve
\begin{verbatim}
## Mutation Strength
- Mutation strength sigma controls mutation scale.

## Self-Adaptation
- Strategy parameters can evolve with candidate solutions.

## Evolution Strategy
- Candidate solutions are generated by mutation.
\end{verbatim}
Retrieval itself does not alter long-term memory. It creates a temporary
working set of models that provides the conceptual context for subsequent
processing.

Semantic retrieval may, for example, be implemented using dense
representations in which queries and stored information are mapped into a
shared vector space \cite{karpukhin2020dpr}.

\section{Memory Management}

\subsection{Add}

The \emph{add} operator extends a mental model with information that is
not yet represented. Its main requirement is novelty. Before adding a
chunk, the manager should determine whether its meaning is already
contained in an existing chunk, even if that knowledge is expressed using
different words.

This semantic novelty test is important for controlling memory growth.
A conventional memory system may store every incoming statement. In 3M,
a new chunk is created only when it increases the represented knowledge
of the model.

Suppose the current model contains
\begin{verbatim}
## Evolution Strategy
- Candidate solutions are generated by mutation.
- Selection favors candidates with better fitness.
\end{verbatim}
A new text states
\begin{quote}
The $(1+1)$-ES generates one offspring from one parent.
\end{quote}
This information is not represented by the existing chunks and is
therefore added:
\begin{verbatim}
## Evolution Strategy
- Candidate solutions are generated by mutation.
- Selection favors candidates with better fitness.
- The (1+1)-ES generates one offspring from one parent.
\end{verbatim}
Thus, addition increases the information content of a model rather than
merely increasing its textual size.

\subsection{Update}

New information frequently does not introduce an entirely new fact but
provides a more accurate, specific, or useful formulation of existing
knowledge. The \emph{update} operator modifies an existing chunk in such
situations.
Updating is preferable to addition when two chunks would otherwise
represent nearly the same knowledge. It allows a model to improve its
precision over time while keeping the representation compact. An update
may add conditions, replace an overly general statement, introduce more
precise terminology, or incorporate additional detail.

The general problem is related to belief-revision research, which studies
principled changes of a knowledge state in response to new information
\cite{alchourron1985logic}.

Suppose memory contains
\begin{verbatim}
- Mutation changes candidate solutions.
\end{verbatim}
and a new source explains that Gaussian mutation is commonly used in
Evolution Strategies. Rather than storing two strongly overlapping
chunks, the manager may update the original chunk:
\begin{verbatim}
- Mutation commonly modifies candidate solutions using
  Gaussian perturbations.
\end{verbatim}
The information content has increased, while the number of chunks remains
unchanged.

\subsection{Merge}

Mental models may be created independently even though they represent
the same underlying concept. This can occur because different sources
use different terminology or because the extraction process initially
fails to recognize a semantic equivalence. The \emph{merge} operator
combines such models.

Merging operates at the model level rather than merely at the chunk
level. It combines descriptions, chunks, aliases, and relations and then
removes redundancies. The operation is particularly useful when knowledge
has been collected over long periods or from heterogeneous sources.

Consider
\begin{verbatim}
## Step Size
- Step size determines mutation magnitude.

## Mutation Strength
- Mutation strength sigma controls mutation size.
\end{verbatim}
Both models describe the same central ES parameter. They can be merged
into
\begin{verbatim}
## Mutation Strength
Aliases: step size

- Mutation strength sigma controls mutation scale.
\end{verbatim}
The alias can be retained for future retrieval, while the knowledge itself
is represented only once.

\subsection{Split}

The opposite problem occurs when one mental model accumulates knowledge
about several concepts that should be represented independently. Such
models become difficult to retrieve and modify because their conceptual
boundaries are unclear. The \emph{split} operator divides an overly broad
model into several more coherent models.

A split can be triggered when groups of chunks within one model have
weak semantic relations or correspond to established sub-concepts.
Relations between the resulting models can be retained so that structural
information is not lost.

Consider
\begin{verbatim}
## ES Parameters
- sigma determines mutation magnitude.
- lambda determines the number of offspring.
- Recombination combines information from parents.
\end{verbatim}
This model mixes mutation strength, population structure, and
recombination. A split produces
\begin{verbatim}
## Mutation Strength
- sigma determines mutation magnitude.

## Population
- lambda determines the number of offspring.

## Recombination
- Recombination combines information from parents.
\end{verbatim}
The resulting models are individually easier to retrieve, extend, and
relate to other concepts.

\subsection{Connect}

Mental models become substantially more useful when their relationships
are represented explicitly. The \emph{connect} operator establishes
semantic, logical, causal, hierarchical, or procedural relations between
models.
Connections allow the memory to represent not only what concepts mean
but also how they interact. They can additionally support retrieval:
accessing one model may lead to related models through its connections.

For example, suppose models for mutation strength and exploration already
exist. The manager can establish
\begin{verbatim}
## Relations
- Mutation Strength -> controls -> Exploration Scale
\end{verbatim}
Further processing may create additional connections such as
\begin{verbatim}
- Selection -> evaluates -> Mutated Offspring
- Self-Adaptation -> modifies -> Mutation Strength
\end{verbatim}
The mental-model memory thereby develops a conceptual network rather than
remaining a flat collection of independent entries.

Related ideas appear in graph-structured retrieval and agent-memory
systems that explicitly connect entities or memory entries
\cite{edge2024graphrag,xu2025amem}.

\subsection{Compress}

As models accumulate information, several chunks may describe different
aspects of the same underlying principle. The \emph{compress} operator
replaces such groups with a smaller representation while attempting to
preserve their relevant information.

Compression differs from pruning. Pruning removes information that is
redundant or no longer useful. Compression instead constructs a new chunk
that summarizes the essential content of several existing chunks.

For example,
\begin{verbatim}
- Small sigma produces small mutations.
- Small mutations search locally.
- Very small sigma can produce little progress.
- Small sigma reduces exploration.
\end{verbatim}
can be compressed to
\begin{verbatim}
- Small sigma favors local search but can cause slow
  progress when it becomes too small.
\end{verbatim}
Compression is particularly important for bounded memory because it
allows knowledge to accumulate without proportional growth in the number
of chunks.

\subsection{Prune}

Mental-model memory can grow even when addition is controlled. Some
chunks become redundant after updates or compression, while others may
become obsolete when a more accurate representation is introduced. The
\emph{prune} operator removes such chunks.
Pruning differs from compression because it does not attempt to derive a
new representation from several chunks. Instead, it removes information
whose relevant content is already represented elsewhere or whose value
has disappeared.

For example, suppose memory contains
\begin{verbatim}
- sigma determines step size.
- Step size depends on sigma.
- sigma determines mutation magnitude.
- Mutation strength controls mutation scale.
\end{verbatim}
The four statements represent essentially the same relationship. The
manager can retain the canonical representation
\begin{verbatim}
- Mutation strength sigma controls mutation scale.
\end{verbatim}
and remove the remaining chunks.

Pruning can also remove temporary hypotheses after they have been
rejected, outdated chunks after an update, or highly specific information
that no longer contributes to the intended purpose of a model. It
therefore provides a mechanism for forgetting in addition to learning.

Selective retention and forgetting have also been explored in long-term
LLM memory systems such as MemoryBank \cite{zhong2023memorybank}.

\section{Cognitive Processing}

\subsection{Conflict Detection}

Information from different sources is not always mutually consistent.
Furthermore, a newly inferred principle may conflict with previously
stored knowledge. The \emph{conflict detection} operator searches for
pairs or groups of chunks that cannot simultaneously be accepted without
additional qualification.

Importantly, an apparent conflict should not immediately cause one chunk
to be deleted. Statements may refer to different conditions, scales, or
contexts. Conflict detection therefore marks the inconsistency as a
reasoning problem for subsequent repair.

Suppose memory contains
\begin{verbatim}
- Large mutation strength improves exploration.
\end{verbatim}
while a new source states
\begin{verbatim}
- Large mutation strength can reduce optimization performance.
\end{verbatim}
The system records
\begin{verbatim}
## Conflict
A: Large mutation strength improves exploration.
B: Large mutation strength can reduce optimization performance.
\end{verbatim}
The conflict indicates that the current representation is insufficiently
precise.

\subsection{Conflict Repair}

The \emph{conflict repair} operator attempts to transform inconsistent
chunks into a coherent representation. Several repair strategies are
possible. One statement may be erroneous, both may apply under different
conditions, or a more general representation may explain why both
observations occur.
Repair is therefore a reasoning operation rather than simple deletion.
Where possible, it should preserve valid information while making the
conditions under which each statement holds explicit.

For the previous example, the manager can distinguish exploration from
optimization performance:
\begin{verbatim}
- Increasing sigma increases exploration.
- Excessively large sigma can cause overshooting and
  reduce optimization performance.
\end{verbatim}
The contradiction disappears because the model now represents the
trade-off between exploration and useful progress.

\subsection{Generalize}

A collection of specific observations may imply a principle that applies
more broadly than any individual observation. The \emph{generalize}
operator searches for such common structure and creates a more general
chunk.
The construction of higher-level knowledge from accumulated memories has
also appeared in reflective agent architectures
\cite{park2023generative}; 3M represents such transformations explicitly
as operators over persistent mental models.

Generalization is a cognitive operation because the resulting knowledge
need not occur explicitly in the input. The operation should therefore
be conservative and ideally accompanied by information about the chunks
from which the generalization was derived.

Suppose memory contains
\begin{verbatim}
- Large sigma can cause overshooting.
- Small sigma can cause slow progress.
- Useful sigma changes during optimization.
\end{verbatim}
A generalization is
\begin{verbatim}
- Effective search requires mutation scale to match
  the current search state.
\end{verbatim}
This principle can subsequently be reused when reasoning about other ES
variants or adaptation mechanisms.

\subsection{Specialize}

While generalization moves toward broader principles, the
\emph{specialize} operator applies general knowledge to a particular
problem, environment, or subclass. Specialization is useful when a
general model provides guidance but additional contextual assumptions
allow a more specific conclusion.

The specialized model should remain connected to the general model. This
makes it possible to distinguish knowledge that applies broadly from
knowledge that applies only under particular conditions.

Starting from
\begin{verbatim}
- Effective search requires mutation scale to match
  the current search state.
\end{verbatim}
the manager may specialize the principle for the Sphere function:
\begin{verbatim}
## Sphere Optimization
- Useful mutation strength generally decreases as the
  search approaches the optimum.
\end{verbatim}
The original principle is retained, while the specialized model captures
its interpretation for a particular optimization landscape.

\subsection{Abstract}

The \emph{abstract} operator creates higher-level mental models from
several related models. While generalization typically creates a broader
chunk from several observations, abstraction operates primarily on the
organization of complete concepts. It identifies a common structure
shared by several mental models and represents this structure as a new
model.

Consider three existing models:
\begin{verbatim}
## Mutation-Strength Adaptation
- sigma can adapt during optimization.

## Covariance Adaptation
- The mutation distribution can adapt its shape.

## Population-Size Adaptation
- Population size can change during optimization.
\end{verbatim}
All three describe mechanisms through which parameters controlling the
search process change during optimization. The abstraction operator may
therefore construct

\begin{verbatim}
## Strategy-Parameter Adaptation

Description:
Adaptation of parameters controlling the search process.

- Evolution Strategies can adapt parameters controlling
  their search behavior.
- Adaptation can affect scale, shape, direction, or
  population structure.

Instances:
- Mutation-strength adaptation
- Covariance adaptation
- Population-size adaptation
\end{verbatim}
The new model does not replace the original models. Instead, it creates a
higher level in the conceptual hierarchy. Retrieval of
\emph{Strategy-Parameter Adaptation} can subsequently provide access to
the more specialized models.

Abstraction is particularly relevant when mental-model memory becomes
large. Rather than retrieving hundreds of individual models, an LLM can
first reason with higher-level abstractions and descend to detailed
models only when necessary.

\subsection{Infer}

The \emph{infer} operator derives knowledge that follows from existing
chunks. Unlike extraction, inference does not require the conclusion to
appear explicitly in an external source. It combines existing knowledge
according to logical, causal, or domain-specific relationships.

Inference can make memory more useful because important consequences of
stored knowledge become explicitly available for later retrieval.
However, inferred chunks should remain distinguishable from directly
observed chunks so that their origin can be inspected or verified.

Suppose memory contains
\begin{verbatim}
- Mutation strength can be inherited.
- Different offspring can have different mutation strengths.
- Selection favors offspring with better fitness.
\end{verbatim}
The manager may infer
\begin{verbatim}
- Selection can indirectly favor mutation strengths that
  generate successful offspring.
\end{verbatim}
This inferred chunk provides a conceptual explanation for why
self-adaptation can work.

\subsection{Analogy}

The \emph{analogy} operator searches for structural similarities between
mental models. Rather than requiring identical concepts, analogy maps
roles and relations from one domain onto another. Such mappings can
support explanation, hypothesis generation, and transfer of reasoning
strategies.

An analogy should normally be represented separately from factual
knowledge because structural similarity does not guarantee that all
properties transfer between domains.

Mutation-strength adaptation can be compared with walking toward a
target:
\begin{verbatim}
Evolution Strategy  <-> Walking
Candidate solution  <-> Current position
Mutation             <-> Step
Mutation strength    <-> Stride length
Optimum              <-> Target
\end{verbatim}
The structural correspondence suggests
\begin{verbatim}
- Mutation-strength adaptation resembles adjusting stride
  length: large steps can be useful far from the target,
  while smaller steps become useful near the target.
\end{verbatim}
The analogy provides an intuitive model of adaptation and may also
suggest questions about when step sizes should increase or decrease.

\subsection{Find Gap}

A useful knowledge system should represent not only what it knows but
also what is missing. The \emph{find gap} operator examines a mental
model for unanswered questions, missing mechanisms, unexplained
relations, or incomplete causal chains.

Knowledge gaps can be used actively. They may trigger retrieval from
external sources, direct the reading of additional documents, generate
research questions, or guide an agent toward an experiment.

Suppose memory contains
\begin{verbatim}
## Mutation Strength
- sigma controls mutation scale.
- Appropriate sigma depends on the search state.
- sigma should therefore be adapted.
\end{verbatim}
The model states that adaptation is necessary but does not explain how
adaptation should be performed. The manager may therefore introduce
\begin{verbatim}
## Knowledge Gaps
- How can the algorithm determine whether sigma should
  increase or decrease?
- Which observations can be used for adaptation?
- How quickly should sigma react to new observations?
\end{verbatim}
The gap is not treated as factual knowledge. Instead, it explicitly
represents an unresolved part of the mental model. A subsequent retrieval
operation may search for models related to success-based adaptation,
self-adaptation, or cumulative step-size adaptation. In this way,
identified gaps can direct future knowledge acquisition.

\subsection{Verify}

Mental models can contain errors originating from unreliable sources,
incorrect extraction, overly strong generalizations, or invalid
inferences. The \emph{verify} operator examines whether a chunk is
supported by the available knowledge and whether its formulation is
sufficiently precise.

Verification may use several sources of evidence. A chunk can be compared
with other mental models, checked against retrieved external information,
tested for logical consistency, or evaluated experimentally when an
appropriate tool is available. Verification can therefore operate at
different levels of confidence.

Consider the stored chunk
\begin{verbatim}
- Increasing sigma improves exploration efficiency.
\end{verbatim}
The statement combines two different claims. A larger $\sigma$ increases
the scale of mutations, but larger mutations do not necessarily improve
useful exploration or optimization performance. Verification may
therefore reformulate the knowledge as
\begin{verbatim}
- Increasing sigma increases mutation scale.
- Whether larger mutations improve exploration efficiency
  depends on the problem and current search state.
\end{verbatim}
Verification is especially important for derived knowledge. Generalized,
inferred, repaired, and analogy-based chunks contain information produced
by cognitive operations of the LLM and should therefore be treated as
claims that may require additional validation.

\section{Implementation and Evaluation}
\label{sec:implementation}

To demonstrate that the proposed operators can be realized in a small
executable system, we implemented a Markdown-native reference manager.
The implementation is one operationalization of 3M rather than a required
execution policy. It focuses on ingesting text into persistent mental-model
memory; alternative controllers may invoke the same operators in different
orders for consolidation, reasoning, auditing, or retrieval.
\footnote{The complete reference implementation, operator prompts, example inputs,
and behavioral tests are available at
\texttt{https://github.com/evolution-strategies/mental-model-management}.}

\subsection{Implementation}

The manager is a single Python process with no third-party runtime
dependencies. It communicates with Ollama through its local HTTP API and was
run on an NVIDIA Spark using \texttt{qwen3.6} at temperature zero.
Each 3M operator has a separate prompt, while a controller chooses the
operators needed for the current input. JSON is used only for transient
controller and operator responses. Persistent knowledge is stored exclusively
as Markdown.

For the ingestion demonstration, the manager uses the outer control cycle

\begin{equation}
T
\xrightarrow{\mathrm{Extract}}
C
\xrightarrow{\mathrm{Retrieve}}
{\cal M}_{C}
\xrightarrow{o_1,\ldots,o_k}
\widetilde{\cal M}
\xrightarrow{\mathrm{Verify}}
{\cal M}',
\end{equation}

where $T$ is the input, $C$ is a transient set of concepts and claims,
${\cal M}_{C}$ is the retrieved memory subset, $\widetilde{\cal M}$ is the
staged state, and ${\cal M}'$ is the committed state. Each $o_i$ is an
operator selected from the 3M vocabulary, for example Add, Update, Connect,
Infer, Conflict Repair, or Abstract. The number $k$ may be zero when no
persistent change is required. Extract and Retrieve do not write memory;
verification and commit are skipped when no file changes.

The implementation imposes several engineering constraints that are not
theoretical requirements of 3M. Each file represents one primary concept,
has a semantic filename, and is limited to 1,500 words by default. Long
multi-concept inputs with empty memory initially produce three to seven
focused models. Operator-specific contracts require, for example, premises
and derived knowledge for Infer, instances for Abstract, relations for
Connect, and a repair audit for Conflict Repair. Proposed changes are
validated before commit; existing files are backed up and replaced
atomically, and the model cannot delete files directly.

\subsection{Cold-Start Demonstration}

The demonstration ingested a 1,577-word introduction to Evolution Strategies
(ES) into an empty memory directory. The manager executed

\begin{center}
\texttt{Extract $\rightarrow$ Retrieve $\rightarrow$ Add $\rightarrow$
Connect $\rightarrow$ Verify}.
\end{center}

Extract produced transient candidate concepts, claims, relations, and
knowledge gaps. Retrieve returned an empty set because no prior memory was
available. The controller consequently selected Add, which created three
staged Markdown files: \emph{Evolution Strategies},
\emph{Covariance Matrix Adaptation}, and
\emph{Search Distribution Adaptation}. Connect then added explicit
relationships among the concepts and their associated mechanisms. Finally,
Verify compared the staged representations with the source input and allowed
the files to be committed.

All controller decisions, operator responses, and Markdown content in the
reported run were generated by the locally executed
\texttt{qwen3.6} model. No generated mental-model file was manually
edited, and no previously constructed mental model was supplied to the empty
test memory.

The committed files contained 471, 260, and 329 words, respectively. Their
combined length was 1,060 words, or 67.2\% of the input length, with a mean
file length of 353.3 words (range 260--471). Across the three files, the
manager stored 24 explicit relations and four knowledge gaps. The reduction
in length should not be interpreted as a compression score, since the output
additionally contains generated relations and knowledge gaps. Rather, the
result shows that the source was transformed into a smaller set of linked,
concept-centered representations instead of being stored as a document-level
summary.

\subsection{Behavioral Evaluation}

We constructed six behavioral test cases with deterministic evaluation
criteria, each executed in a fresh temporary Markdown directory. The cases
specify expected operators, file-count bounds, required sections or terms,
verification status, and bounds on redundancy. They allow multiple valid
wordings and model organizations rather than requiring exact text matches.

Redundancy is approximated by the maximum pairwise token-set Jaccard
similarity $J$ between mental-model files. In the following,
\emph{Changed} denotes the number of persistent Markdown files modified
during a case, while \emph{New} denotes newly created files.

\begin{center}
\begin{tabular}{p{0.3\linewidth}p{0.25\linewidth}rrr}
\hline
\textbf{Case} & \textbf{Selected operators} &
\textbf{Changed} & \textbf{New} & \textbf{Max $J$} \\
\hline
Cold-start decomposition &
Add, Connect & 3 & 3 & .230 \\
Incremental success rule &
Update, Connect & 3 & 1 & .300 \\
Redundant paraphrase &
Prune & 0 & 0 & .000 \\
Conflicting claim &
Conflict Detection, Conflict Repair & 2 & 0 & .219 \\
Self-adaptation reasoning &
Infer, Connect & 4 & 1 & .417 \\
Adaptation hierarchy &
Abstract, Connect & 4 & 2 & .478 \\
\hline
\end{tabular}
\end{center}

With \texttt{qwen3.6}, all six cases satisfied the 67 predefined
structural and behavioral checks. The runs produced 16 changed-file events
and seven new files through 11 operator applications spanning eight distinct
operator types. Maximum within-case Jaccard overlap ranged from .000 to .478,
with a mean of .274, and remained below the case-specific bounds in every
run.

More importantly, the cases demonstrate qualitatively different memory
transformations. Redundant input left persistent memory unchanged, while new
information was integrated through Update. A conflicting claim triggered
explicit detection and repair rather than simple addition. Stored premises
triggered Infer, and several specialized models triggered Abstract. Thus,
the operators correspond to observably different transformations of the
persistent representation rather than merely serving as descriptive labels.

These results demonstrate behavioral coverage rather than general model
quality. The cases were used during development, only one model run was
recorded per case, and no confidence intervals can therefore be reported.
Verification also uses the same language model that proposes changes, and
token overlap is only a coarse redundancy measure. The evaluation should
accordingly be interpreted as an executable proof of concept and regression
suite, not as a held-out comparison or evidence of factual reliability.
Broader evaluation should include unseen domains, repeated runs, independent
annotations, and verification by models or humans not involved in generation.

\section{Discussion and Limitations}

Mental Model Management extends external LLM memory from passive storage
toward an actively maintained conceptual representation. While retrieval
primarily determines which stored information should be made available,
3M additionally determines how persistent knowledge should be updated,
connected, compressed, or reorganized.

A key property is that processing additional text does not necessarily
increase memory size. Redundant information may leave memory unchanged,
while updates, merging, compression, and pruning can reduce redundancy.
Cognitive operations such as inference, abstraction, analogy, and conflict
repair can further improve the representation without adding new external
text.

The current implementation nevertheless depends strongly on the
underlying LLM. Operator selection, semantic comparison, inference, and
verification may all introduce errors. Verification is also not fully
independent when the same model proposes and evaluates a transformation.
Future work should therefore consider independent verification, external
sources, different controller policies, and larger-scale experiments.

The present evaluation is limited to one technical domain and one model.
Further studies should investigate other domains, models, repeated runs,
and long-term memory development. Larger memories may additionally require
embeddings, indexes, or graph-based retrieval while preserving the
concept-centered representation.

\section{Conclusion}

We introduced Mental Model Management (3M), an operator-based framework
for maintaining persistent conceptual knowledge for large language
models. Knowledge is represented as evolving mental models composed of
compact chunks rather than primarily as stored text passages.

The framework combines knowledge acquisition, memory management, and
cognitive processing. A minimal Markdown-based implementation shows that
the proposed operators can produce different transformations of persistent
memory, including refinement, compression, conflict repair, inference,
and abstraction.

The central idea is that processing more information should not merely
increase stored text. It should improve how knowledge is represented.
3M therefore complements retrieval with an explicit mechanism for
maintaining increasingly compact, coherent, and useful conceptual memory.


\begin{thebibliography}{99}

\bibitem{lewis2020rag}
P. Lewis, E. Perez, A. Piktus, F. Petroni, V. Karpukhin,
N. Goyal, H. K{\"u}ttler, M. Lewis, W.-t. Yih,
T. Rockt{\"a}schel, S. Riedel, and D. Kiela,
Retrieval-Augmented Generation for Knowledge-Intensive NLP Tasks,
\emph{Advances in Neural Information Processing Systems}, 33,
9459--9474, 2020.

\bibitem{karpukhin2020dpr}
V. Karpukhin, B. O\u{g}uz, S. Min, P. Lewis, L. Wu,
S. Edunov, D. Chen, and W.-t. Yih,
Dense Passage Retrieval for Open-Domain Question Answering,
\emph{Proceedings of EMNLP}, 6769--6781, 2020.

\bibitem{park2023generative}
J. S. Park, J. C. O'Brien, C. J. Cai, M. R. Morris,
P. Liang, and M. S. Bernstein,
Generative Agents: Interactive Simulacra of Human Behavior,
\emph{Proceedings of the ACM Symposium on User Interface
Software and Technology (UIST)}, 2023.

\bibitem{packer2023memgpt}
C. Packer, S. Wooders, K. Lin, V. Fang, S. G. Patil,
I. Stoica, and J. E. Gonzalez,
MemGPT: Towards LLMs as Operating Systems,
arXiv:2310.08560, 2023.

\bibitem{zhong2023memorybank}
W. Zhong, L. Guo, Q. Gao, H. Ye, and Y. Wang,
MemoryBank: Enhancing Large Language Models with Long-Term Memory,
\emph{Proceedings of the AAAI Conference on Artificial Intelligence},
38(17), 19724--19731, 2024.

\bibitem{xu2025amem}
W. Xu, Z. Liang, K. Mei, H. Gao, J. Tan, and Y. Zhang,
A-MEM: Agentic Memory for LLM Agents,
arXiv:2502.12110, 2025.

\bibitem{yu2026agentic}
Y. Yu, L. Yao, Y. Xie, Q. Tan, J. Feng, Y. Li, and L. Wu,
Agentic Memory: Learning Unified Long-Term and Short-Term Memory
Management for Large Language Model Agents,
arXiv:2601.01885, 2026.

\bibitem{maharana2024locomo}
A. Maharana, D.-H. Lee, S. Tulyakov, M. Bansal,
F. Barbieri, and Y. Fang,
Evaluating Very Long-Term Conversational Memory of LLM Agents,
\emph{Proceedings of the 62nd Annual Meeting of the Association
for Computational Linguistics}, 2024.

\bibitem{edge2024graphrag}
D. Edge, H. Trinh, N. Cheng, J. Bradley, A. Chao,
A. Mody, S. Truitt, and J. Larson,
From Local to Global: A Graph RAG Approach to Query-Focused
Summarization,
arXiv:2404.16130, 2024.

\bibitem{zhang2024memorysurvey}
Z. Zhang, X. Bo, C. Ma, R. Li, X. Chen, Q. Dai,
J. Zhu, Z. Dong, and J.-R. Wen,
A Survey on the Memory Mechanism of Large Language Model Based Agents,
arXiv:2404.13501, 2024.

\bibitem{johnsonlaird1983mental}
P. N. Johnson-Laird,
\emph{Mental Models: Towards a Cognitive Science of Language,
Inference, and Consciousness},
Cambridge University Press, Cambridge, 1983.

\bibitem{alchourron1985logic}
C. E. Alchourr{\'o}n, P. G{\"a}rdenfors, and D. Makinson,
On the Logic of Theory Change: Partial Meet Contraction and
Revision Functions,
\emph{The Journal of Symbolic Logic}, 50(2), 510--530, 1985.

\bibitem{hansenostermeier2001}
N. Hansen and A. Ostermeier,
Completely Derandomized Self-Adaptation in Evolution Strategies,
\emph{Evolutionary Computation}, 9(2), 159--195, 2001.

\bibitem{hansen2016cma}
N. Hansen,
The CMA Evolution Strategy: A Tutorial,
arXiv:1604.00772, 2016.

\end{thebibliography}
\end{document}